\documentclass[conference]{IEEEtran}
\IEEEoverridecommandlockouts
\usepackage{cite}
\usepackage{amsmath,amssymb,amsfonts}
\usepackage{algorithmic}
\usepackage{graphicx}
\usepackage{textcomp}
\usepackage{xcolor}
\usepackage{url}
\def\BibTeX{{\rm B\kern-.05em{\sc i\kern-.025em b}\kern-.08em
    T\kern-.1667em\lower.7ex\hbox{E}\kern-.125emX}}

\ifCLASSOPTIONcompsoc
    \usepackage[caption=false,font=normalsize,labelfont=sf,textfont=sf]{subfig}
\else
    \usepackage[caption=false,font=footnotesize]{subfig}
\fi

\begin{document}
\title{Semantic Segmentation of Node and Edge Diagrams for Assistive Technology \thanks{We acknowledge the support of the Natural Sciences and Engineering Research Council of Canada (NSERC), funding reference number 2020-04401.}}

\author{\IEEEauthorblockN{Michael Cormier}
\IEEEauthorblockA{\textit{Dept. of Mathematics and Comp. Sci.} \\
\textit{Mount Allison University}\\
Sackville, NB, Canada \\
micormier@mta.ca}
\and
\IEEEauthorblockN{Yichun Zhao}
\IEEEauthorblockA{\textit{Department of Computer Science} \\
\textit{University of Victoria}\\
Victoria, BC, Canada\\
yichunzhao@uvic.ca}
\and
\IEEEauthorblockN{Laura Paul}
\IEEEauthorblockA{\textit{Department of Computer Science} \\
\textit{University of Victoria}\\
Victoria, BC, Canada \\
lepaul@uvic.ca}
\and[\hfill\mbox{}\par\mbox{}\hfill]
\IEEEauthorblockN{Cameron Swift}
\IEEEauthorblockA{\textit{Dept. of Mathematics and Comp. Sci.} \\
\textit{Mount Allison University}\\
Sackville, NB, Canada \\
caswift@mta.ca}
\and
\IEEEauthorblockN{Duc Tri Dang}
\IEEEauthorblockA{\textit{Dept. of Mathematics and Comp. Sci.} \\
\textit{Mount Allison University}\\
Sackville, NB, Canada \\
ddang@mta.ca}
\and
\IEEEauthorblockN{Miguel Nacenta}
\IEEEauthorblockA{\textit{Department of Computer Science} \\
\textit{University of Victoria}\\
Victoria, BC, Canada \\
nacenta@uvic.ca}
}

\maketitle

\begin{abstract}
In this paper, we present a novel set of related models for semantic segmentation of node-link diagrams. These diagrams are frequently used to represent mathematical graphs, relationships between concepts, and flowcharts. Such diagrams are difficult to access non-visually; while some assistive interfaces have been designed for node-link diagrams, they rely upon a machine-readable representation of the diagram, whereas such diagrams will generally be made available as bitmap images. Our compact deep learning models show excellent quantitative and qualitative performance on a large synthetic dataset of node-link diagrams, reaching per-pixel accuracy over 93\%.
\end{abstract}

\section{Introduction}
\label{sec:intro}
When people encounter information in documents such as when studying textbooks for a
university course, when browsing a business report or when reading a scientific article, they often come across visual representations of conceptual information. These representations, often referred to as \emph{diagrams}, can be of many different types, but are generally acknowledged to be useful to convey complex information~\cite{larkinWhyDiagramSometimes1987} and to complement prose explanations~\cite{Butcher_2006}. A common and versatile variant of such diagrams is the node-link diagram, sometimes referred to as the ``boxes and arrows'' diagram. These often represent mathematical graph structures or datasets where nodes correspond to topics or objects, and links can represent unidirectional or bidirectional relationships between nodes. 

Unfortunately, this kind of diagrams is difficult to access non-visually, which negatively affects blind or low-vision (BLV) individuals. Until recently, the most common workarounds for BLV people were to access textual descriptions of the diagram or to ask sighted people to interpret the diagram on their behalf. Both have obvious drawbacks: pre-made textual descriptions are unlikely to match the need for an specific type of information required (\textit{e.g.}, we might want an overview and get a listing of all nodes and how they are connected, or vice versa), and asking others, even through a service like BeMyEyes~\cite{beMyEyes}, has time and social costs.

More recently, the availability of large language models (LLMs) and multi-modal large models (MLMs) has made it possible to upload a picture of a diagram and ask questions to the model about it. Although this is certainly more versatile than the previous approaches, it also introduces some challenges and difficulties of cognitive, educational and communication nature. For example, diagrams support understanding information through spatial relationships are difficult and tedious to communicate through text~\cite{larkinWhyDiagramSometimes1987}; direct interaction with spatial information might be a crucial component in the memorability and cognitive processing of the information~\cite{verhoevenCognitiveLoadInteractive2009}; and it might become difficult for a person to communicate with others if they cannot refer to specific elements, patterns or relationships in spatial ways (\textit{e.g.}, ``that group of tightly related nodes on the bottom left'').

Several research groups have proposed interfaces for the BLV community that enable users to spatially explore diagrams, circumventing most of the human-centered issues described above (\textit{e.g.,}~\cite{zhaoTADAMakingNodelink2023,petrieTeDUBSystemPresenting2002,balikGSKUniversallyAccessible2013}). However, these interfaces typically assume as a starting point a semantically annotated data structure, rather than just an image. It might be possible to use LLMs and MLMs to create such data structures, but this comes with additional concerns: it will likely require uploading information to a central server with sufficient memory and processing power to translate an image into a graph data structure, exposing potentially private information from the user to the network and making the translation unavailable when there is not a sufficiently broadband connection between the user's device and the server. Additionally, there are widespread societal concerns that use of LLMs and MLMs for relatively simple problems has important disadvantages in the use of energy and water, leading to environmental concerns.

The translation of images or photos of diagrams to graph data structures is a good candidate for a solution that uses a smaller and more energy-efficient specialized computer vision model that can run on portable devices, therefore avoiding making private data vulnerable to unauthorized access in servers or during transmission. This is because the problem is unlikely to evolve and require outside information and because the structure of the problem is likely to benefit from hybrid machine learning approaches that include human-provided knowledge, in contrast with a brute-force, energy-hungry approach.

This paper provides a major step toward an integrated pipeline for parsing node-and-edge diagrams. The models presented here are compact and robust to expected flaws in rendered images of graphs of this type, and provides a highly accurate semantic segmentation to support later phases of the pipeline. In isolation, the semantic segmentation of diagrams that we propose may also support other forms of assistive interfaces for diagram accessibility.

\section{Related Work}
\label{sec:related}
A broader body of work exists regarding diagram parsing, mostly interpreting charts (e.g., bar charts, line charts, pie charts, etc.) to: produce text descriptions~\cite{balajiChartTextFullyAutomated2018, bhushanBlockDiagramtoTextUnderstanding2022, huangSystemUnderstandingImaged2007, zhuAutoChartDatasetCharttoText2021}; extract data~\cite{daiChartDecoderGenerating2018, gaoViewVisualInformation2012, jungChartSenseInteractiveData2017, liuDataExtractionCharts2019, siegelFigureSeerParsingResultFigures2016, royDiag2graphRepresentingDeep2020a, bhushanBlockDiagramtoTextUnderstanding2022, schaferArrowRCNNFlowchart2019, schaferArrowRCNNHandwritten2021, schaferDiagramNetHandDrawnDiagram2021}; generate some machine-readable output, such as XML, for re-designing diagrams~\cite{pocoReverseEngineeringVisualizations2017, huangSystemUnderstandingImaged2007, savvaReVisionAutomatedClassification2011, zhutianAutomatedInfographicDesign2019}; or answer questions based on the content of diagrams~\cite{kimDynamicGraphGeneration2018a, kimTextbookQuestionAnswering2019, kembhaviDiagramWorthDozen2016, gomez-perezISAAQMasteringTextbook2020, huangGeoSQABenchmarkScenariobased2019, wangCoGDQAChainofGuidingLearning2024, maDiagramPerceptionNetworks2024}.
Only a few, however, parse node-edge diagrams~\cite{royDiag2graphRepresentingDeep2020a, bhushanBlockDiagramtoTextUnderstanding2022, julca-aguilarSymbolDetectionOnline2018, schaferArrowRCNNFlowchart2019, schaferArrowRCNNHandwritten2021, schaferDiagramNetHandDrawnDiagram2021}, and even fewer consider accessibility~\cite{choiVisualizingNonVisualEnabling2019, kimExploringChartQuestion2023}. In this section, we motivate the utility of a system for producing machine-readable representations of node-edge diagrams from an accessibility lens, and discuss previous approaches to parsing node-edge diagrams in specific.

\subsection{Diagram Accessibility}
Node-edge diagrams can be pervasive and powerful tools, capable of showing complex relationships in a simple, understandable manner for sighted users. Previous research has shown, however, that node-edge diagrams are often inaccessible to BLV users 
and can require extensive additional effort for them to understand~\cite{zhaoTADAMakingNodelink2023, choiVisualizingNonVisualEnabling2019, kimExploringChartQuestion2023}. Not only may BLV users miss the existence of diagrams entirely or lack access to visual information within the diagram; current solutions usually provide textual summaries that may be missing information and do not let users control the level of detail (\textit{i.e.}, an overview versus an in-depth description)~\cite{zhaoTADAMakingNodelink2024, choiVisualizingNonVisualEnabling2019}. This forces users to spend extra effort and time parsing the diagram or seeking out other resources~\cite{zhaoTADAMakingNodelink2024, choiVisualizingNonVisualEnabling2019}. Other options may involve extensive cognitive effort to memorize and interpret large data summaries, or incur social costs by asking (``bothering'') sighted people to help the BLV user understand the diagram~\cite{zhaoTADAMakingNodelink2024, choiVisualizingNonVisualEnabling2019}.

Question-answering systems have been explored as a potential solution to this problem, with one study reporting that these systems could provide users with more autonomy over the information they receive in a textual summary, limiting the need to seek out external resources~\cite{kimExploringChartQuestion2023}. However, these systems do not enable users to engage with spatial information encoded within diagrams that may support their understanding~\cite{zhaoTADAMakingNodelink2024}. 
In addition, existing question-answering systems may provide inaccurate representations of the data within a diagram when responding to queries --- state-of-the-art models achieve, at best, approximately 80\% accuracy for answering queries when interpreting diagrams or charts with text, and lower accuracy when interpreting diagrams alone~\cite{kimDynamicGraphGeneration2018a, kimTextbookQuestionAnswering2019, kembhaviDiagramWorthDozen2016, gomez-perezISAAQMasteringTextbook2020, huangGeoSQABenchmarkScenariobased2019, wangCoGDQAChainofGuidingLearning2024, maDiagramPerceptionNetworks2024}.

Tools that can assist BLV people with getting more effective and comprehensive access to diagrammatic information exist; examples include the Audiograf~\cite{kennelAudiografDiagramreaderBlind1996}, TeDUB~\cite{horstmannAutomatedInterpretationAccessible2004, kingPresentingUMLSoftware2004, petrieProvidingInteractiveAccess2006, petrieTeDUBSystemPresenting2002}, GSK~\cite{balikGSKUniversallyAccessible2013, balikIncludingBlindPeople2014}, PLUMB~\cite{calderPLUMBInterfaceUsers2006, cohenPLUMBDisplayingGraphs2005, cohenTeachingGraphsVisually2006, cohenUsingAudioInterface2006}, and TADA~\cite{zhaoTADAMakingNodelink2024} systems. The TADA system in particular provides users with greater access and autonomy when interacting with node-edge diagrams, utilizing proprioception to maintain spatial representations. These systems, however, require a machine-readable encoding (such as an XML, GraphML, or DOT file) of the diagram to render a representation of the diagram. Currently, this severely limits their application in the real-world, but a system capable of generating a machine-readable representation of node-edge diagrams from rasterized images can support these systems becoming more available. A machine learning system that can accurately identify the different components of a node-edge diagram within a rasterized image, such as the one presented in this paper, works towards closing this gap.

\subsection{Parsing Node-Edge Diagrams}
Previous work uses object detection to find instances of nodes and edges in a chart, followed by applying a grammar to reconstruct these components into a meaningful output (\textit{e.g.}, text, knowledge graphs).
The majority build on a paper by Julca-Aguilar and Hirata~\cite{julca-aguilarSymbolDetectionOnline2018}, who applied Faster R-CNN to node detection in handwritten flowcharts and mathematical expressions, finding  convolutional neural networks (CNNs) to be an effective and accurate method for node detection.

The Arrow R-CNN system~\cite{schaferArrowRCNNFlowchart2019, schaferArrowRCNNHandwritten2021} for parsing handwritten flowcharts applies Faster R-CNN with added feature pyramid network (FPN) and region of interest (ROI) alignment to detect nodes. Edges are detected using an additional regression step, integrating a structural grammar to improve arrow detection. The DiagramNet system~\cite{schaferDiagramNetHandDrawnDiagram2021}, also for parsing handwritten flowcharts, similarly uses Faster R-CNN with FPN to detect nodes, but uses a MobileNet-inspired network to detect/predict edges to improve on Arrow R-CNN's accuracy.
And, finally, the BloSum framework~\cite{bhushanBlockDiagramtoTextUnderstanding2022} is designed to create natural language descriptions of block diagrams, with an intermediate `data' stage between the image input and the text output. In the image to data stage, the system uses Faster R-CNN to detect nodes, Hough line detection for edges, and EasyOCR for text. Nodes and text are masked out of the image before line detection to improve the accuracy of the process.

Roy et al.~\cite{royDiag2graphRepresentingDeep2020a} are the exception to using Faster R-CNN. Their Diag2Graph system generates knowledge graphs from deep-learning architecture diagrams (a particular type of node-edge graph), using iterative region growing to find nodes and the EAST text detector followed by OCR for text. Like BloSum, nodes and text are masked out of the image before performing Hough line detection for edges.

Our semantic segmentation system presents a complementary approach to the object-detection methods used in existing systems. It is also intended to be flexible with respect to the library of symbols used for nodes; existing object-detection methods also classify nodes according to shape. A sufficiently accurate semantic segmentation provides valuable insights into diagram structure, and can be accomplished with the compact feed-forward models described in this paper. 

\section{Dataset}
\label{sec:dataset}
\subsection{Synthetic graphs}
Our dataset was generated procedurally by using the Python library Graphviz, with diagrams constructed programatically to achieve visual and structural diversity. Each graph consists of between 2 and 15 nodes, with edges generated probabilistically to create a range of sparse to dense connectivity patterns while avoiding redundant connections between the same node pairs. 

Nodes are assigned randomly selected visual attributes for shape, outline colour, fill colour, font, font colour and text content. Node shapes are sampled from a large predefined set of shapes supported by Graphviz (\textit{e.g.}, rectangles, ellipses, polygons, stars, folders, and components). Node labels consist of randomly generated alphanumeric strings with lengths biased toward typical English word lengths. Fonts are sampled from a small set of common serif and sans-serif fonts, and colours are generated uniformly at random in RGB space. 

Edges are generated by randomly selecting start and end nodes and choosing whether to create an edge between them, with a higher probability of creating an edge than not. Directedness is varied by sampling from \texttt{forward}, \texttt{back}, \texttt{both}, and \texttt{none}. Edge thickness is drawn from a continuous range. Edge labels are generated in the same manner as node labels, with independent random text content, font colour, and thickness. This process results in graphs that vary widely in layout, node density, edge density, and stylistic conventions.

\begin{figure}
    \centering
    \includegraphics[width=0.9\linewidth]{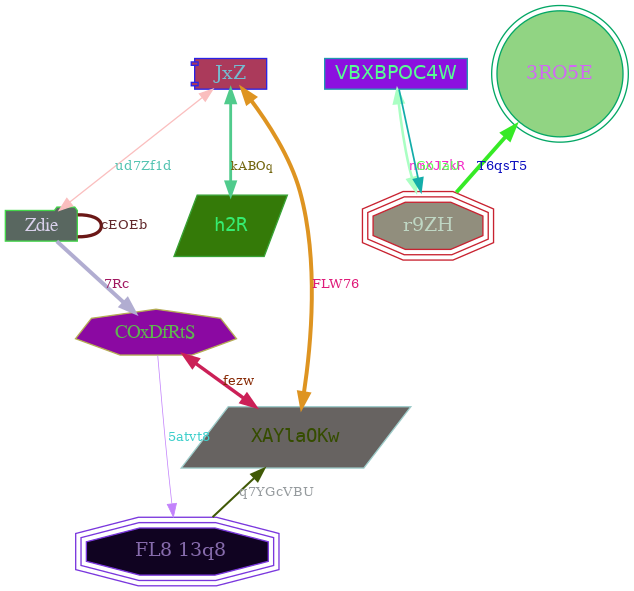}
    \caption{Example graph selected from the test set, showing a variety of node styles.}
    \label{fig:example_graph}
\end{figure}

All graphs are laid out using Graphviz’s \texttt{neato} engine with multidimensional scaling, spline-based edges, and overlap removal enabled. This produces non-grid-based layouts with irregular spacing and edge routing. Figure \ref{fig:example_graph} shows an example graph from the test set. Note that there is considerable variety in the node shapes and styles.

Ground truth data was generated by rendering separate versions of the graphs with colors set to encode both instance and semantic labels. 
For each diagram, four renderings were produced: 1) a visually realistic test image; 2) a semantic label image; 3) an instance-coded label image; and, 4) a text-only label image used to recover accurate text masks.
For the experiments described in this work, only the semantic information was used. 
The label colors were generated as follows (using a range of $[0, 255]$ to represent each color channel, in RGB order), where $i$ is the instance number:
\begin{itemize}
    \item Node border: $[0, 64, 8i]$
    \item Node interior: $[0, 128, 8i]$
    \item Edge: $[0, 192, 8i]$
    \item Text: $[128, 0, 8i]$
    \item Background: $[255, 255, 255]$ (assumed to consist of a single instance)
\end{itemize}

This allows for easy extension to other classes in future work and creates human-readable ground truth label images. For the purposes of semantic segmentation, we merge all instances into a single instance 0. 

To ensure accurate labeling at the pixel level, antialiasing was disabled while rendering these graphs. This was effective for every class except text, which used a separate rendering engine for which antialiasing could not be easily disabled. To label text, therefore, a third image was rendered in which all colors other than that of text was set to white. This allowed the reconstruction of the original color for all pixels corresponding to part of the text despite the use of antialiasing. All pixels with some text component were considered to be a part of the ground truth mask for text regions.

\subsection{Data Augmentation}
Data augmentation is an asset for many computer vision applications. In the case of generated images of graphs, typical sources of noise such as rotation and blurring are not applicable; the image formation process generally does not feature a camera which may be slightly out of focus or slightly tilted. There are, however, other sources of noise which may occur when working with generated images. In particular, images may be aggressively compressed using lossy formats. To perform experiments with degraded images, we have augmented our dataset by compressing images in the JPEG format. While PNG format is more appropriate for this type of diagram, it is not unlikely that diagrams would be encountered where the JPEG format has been used. The conversion was performed on images from the training, validation, and test sets using ImageMagick 6.9.10-23 and a quality setting of 50. New patches were selected from each modified image to generate the training, validation and test sets of image patches.

\section{Models}
\label{sec:models}
\begin{figure}
    \centering
    \includegraphics[width=0.9\linewidth]{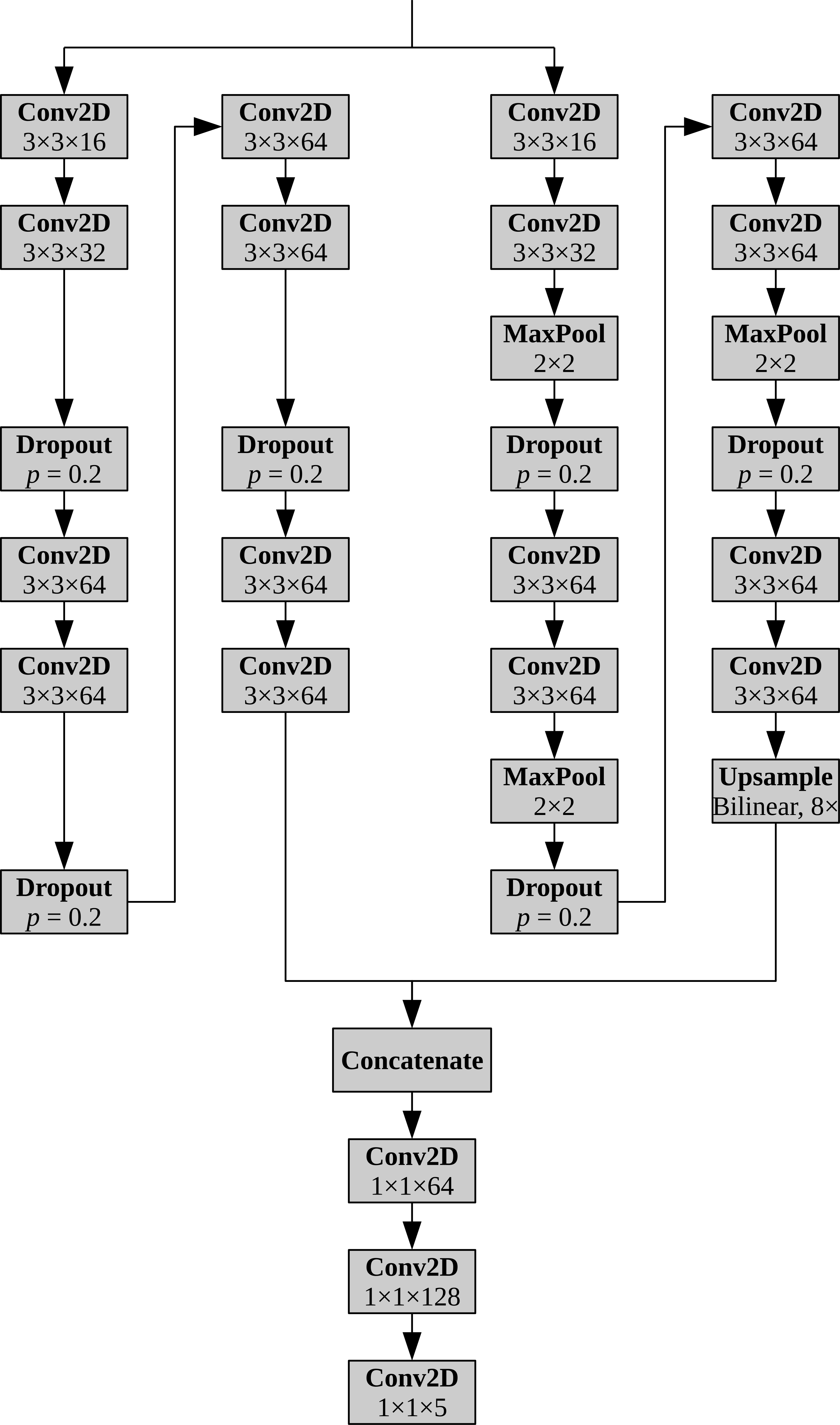}
    \caption{Complete network, showing the fine branch of the network on the left and the coarse branch on the right.}
    \label{fig:fullmodel}
\end{figure}

Our model is a simple and relatively compact fully convolutional neural network with two branches, one of which uses max-pooling to expand the receptive fields of the filters and the other of which has no pooling in order to capture fine details. Dropout layers are used to prevent overfitting. The branches are merged by concatenating the features from each branch, and the final classification of pixels is performed using a series of $1 \times 1$ convolutional layers. All of the convolutional layers use a ReLU activation function. Figure \ref{fig:fullmodel} shows the structure of the complete network. 

The use of a fully convolutional architecture facilitates the application of the model to images of arbitrary size. Without fully connected layers, there are no layers which assume a specific total size; this allows the network to be applied to real images, which occur in a wide range of sizes. For practical purposes, it is more convenient to use a standardized image patch for training and testing, to allow easy batching of samples. In our experiments, a $128 \times 128$ patch is used. These properties are not strictly required to apply the algorithm to large images, as these images could be broken down into patches of a specified size, but it is convenient to have the ability to directly apply the model to images of any size.

To fully evaluate the design of our model, we have tested several variations on the design. Compared to the version shown in Figure \ref{fig:fullmodel}, these variations use exclusively the fine branch and exclusively the coarse branch; the models that use the fine branch were also reduced in depth from eight $3 \times 3$ convolutional layers to four and two convolutional layers and retrained to characterize the performance of the model as its size changes.

\section{Results}
\label{sec:results}
\subsection{Training}
The model was trained on our augmented training set of 5000 patches, each 128-by-128 pixels, using a batch size of 64 and a fixed learning rate of $1 \times 10^{-6}$; the validation and test sets each contain 1000 patches. The loss function was focal loss \cite{lin2017focal}, using the PyTorch implementation of Adeel Hassan\footnote{\url{https://github.com/AdeelH/pytorch-multi-class-focal-loss}}. For training purposes, we used class weights derived from the observed frequencies of classes in the training dataset, and $\gamma=2$. 

\begin{figure*}[!t]
    \centering
    \subfloat[Loss]{\includegraphics[width=3.5in]{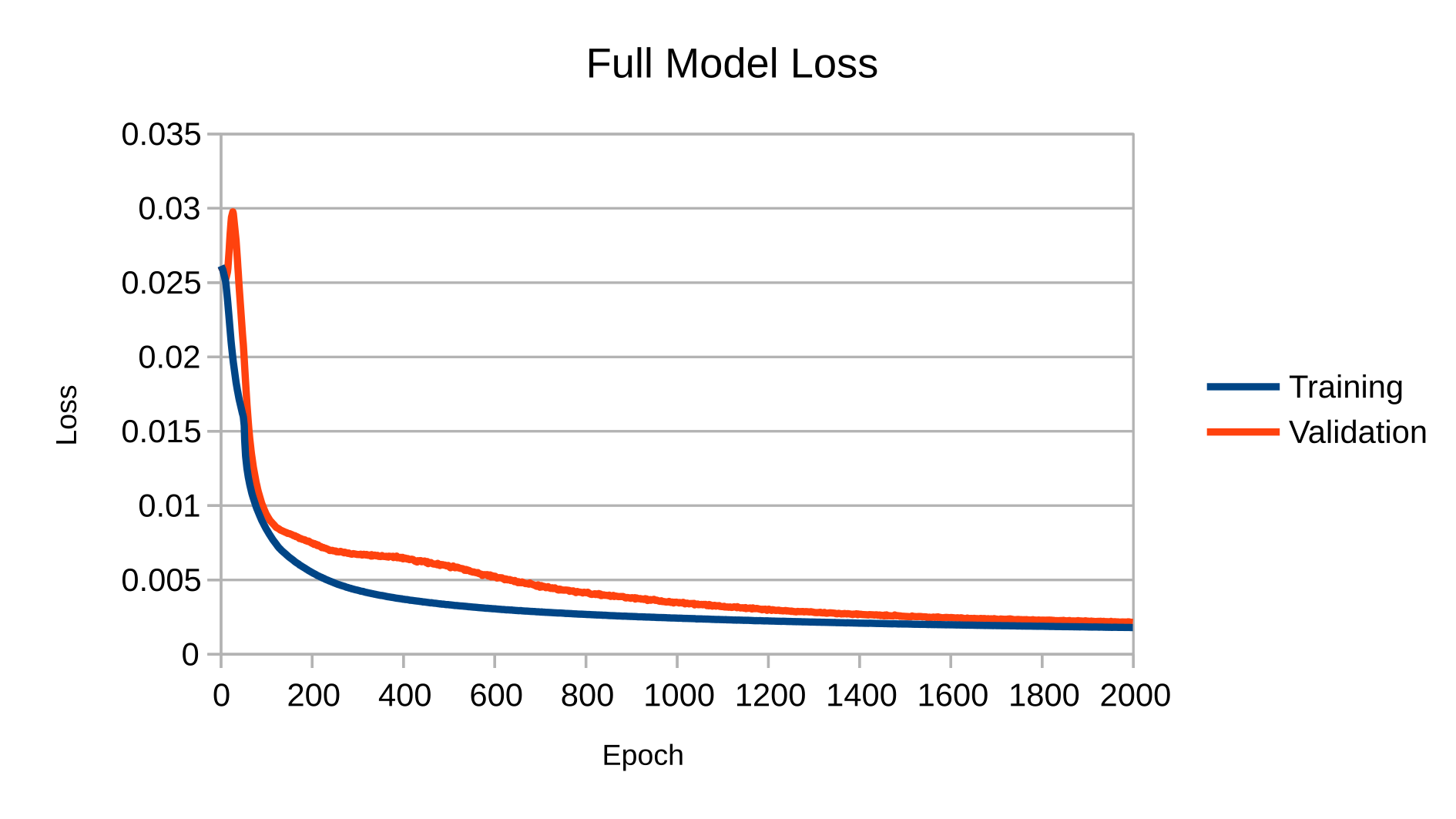}}
    \label{fig:full_training_graphs:loss}
    \hfill
    \subfloat[Accuracy]{\includegraphics[width=3.5in]{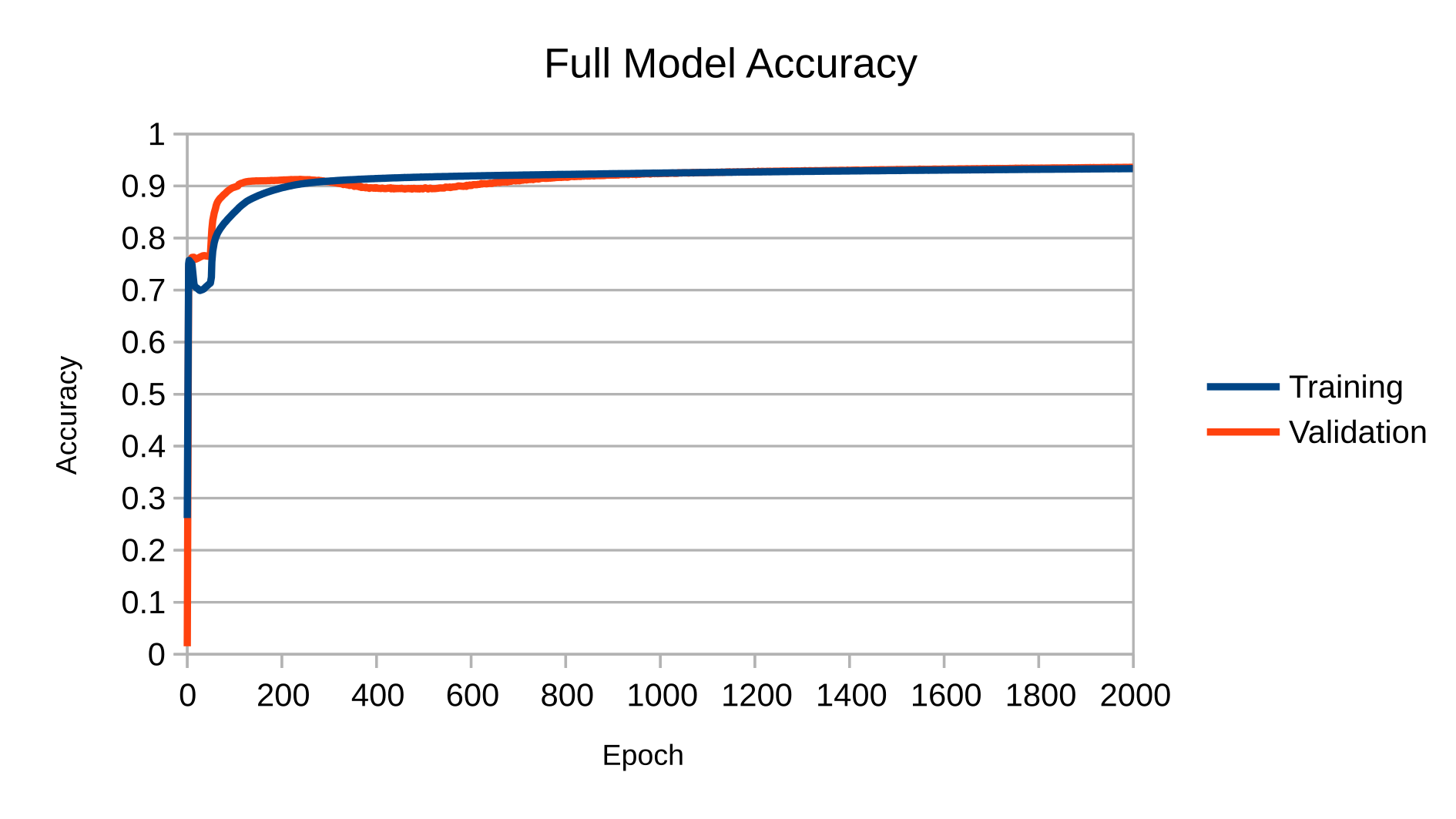}}
    \label{fig:full_training_graphs:accuracy}
    \caption{Training graphs showing weighted focal loss and accuracy for the full model, trained on the augmented dataset.}
    \label{fig:full_training_graphs}
\end{figure*}

Figure \ref{fig:full_training_graphs} shows training graphs for accuracy and average loss on the training and validation sets. These graphs show good convergence to a highly effective pixelwise classifier, with some unevenness in the very early stages of training.

\section{Patches}
\begin{figure*}[!t]
    \centering
    \begin{tabular}{ccccccc}
         Image & Ground Truth & Full & Coarse & Fine (8) & Fine (4) & Fine (2)\\
         \frame{\includegraphics[width=0.8in]{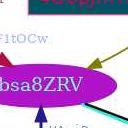}} & \frame{\includegraphics[width=0.8in]{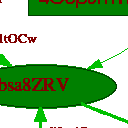}} & \frame{\includegraphics[width=0.8in]{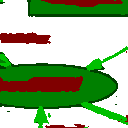}} &
         \frame{\includegraphics[width=0.8in]{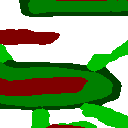}} &
         \frame{\includegraphics[width=0.8in]{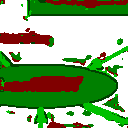}} &
         \frame{\includegraphics[width=0.8in]{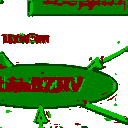}} &
         \frame{\includegraphics[width=0.8in]{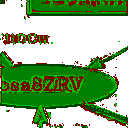}} \\
         \frame{\includegraphics[width=0.8in]{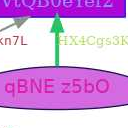}} & \frame{\includegraphics[width=0.8in]{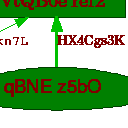}} & \frame{\includegraphics[width=0.8in]{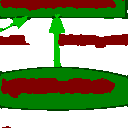}} &
         \frame{\includegraphics[width=0.8in]{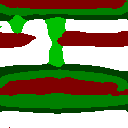}} &
         \frame{\includegraphics[width=0.8in]{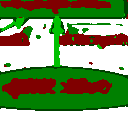}} &
         \frame{\includegraphics[width=0.8in]{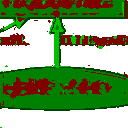}} &
         \frame{\includegraphics[width=0.8in]{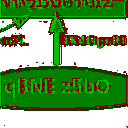}} \\
         \frame{\includegraphics[width=0.8in]{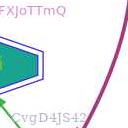}} & \frame{\includegraphics[width=0.8in]{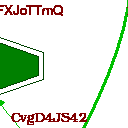}} & \frame{\includegraphics[width=0.8in]{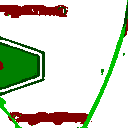}} &
         \frame{\includegraphics[width=0.8in]{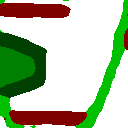}} &
         \frame{\includegraphics[width=0.8in]{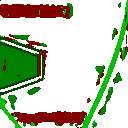}} &
         \frame{\includegraphics[width=0.8in]{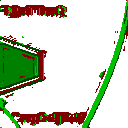}} &
         \frame{\includegraphics[width=0.8in]{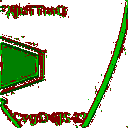}} \\
    \end{tabular}
    \caption{Examples of segmented patches, 128-by-128 pixels, from the augmented test set. The numbers given with the ``Fine'' models represent the number of $3 \times 3$ convolutional layers before the final classification module.}
    \label{fig:examples_full_patches_aug}
\end{figure*}

When trained on the augmented dataset, the full model achieves a pixel-wise accuracy of 93.7\% on the test dataset. Figure \ref{fig:examples_full_patches_aug} shows examples of 128-by-128 pixel patches from the generated graphs in the test set, the corresponding ground truth segmentations, and the segmentations generated by the full model, as well as several variant models which are discussed below. We selected these patches, which are typical examples of the performance on degraded images, specifically to show the robustness of the model with respect to compression artifacts. The compression artifacts have little impact on segmentation quality, although a few isolated pixels of the background are misclassified in these examples.

An examination of the results of classification on patches from the test set demonstrates that the high accuracy values correspond to high-quality semantic segmentations. Not only is the accuracy high, the errors that are observed tend to be qualitatively reasonable; there is, for example, a certain amount of widening of edge and text regions that will be counted as erroneous by our accuracy and loss measurements, but which is perceptually reasonable and unlikely to cause practical problems in use. These examples of the performance of our model therefore provide additional evidence to support its practical utility. 

\subsection{Whole Images}
\begin{figure}
    \centering
    \includegraphics[width=0.9\linewidth]{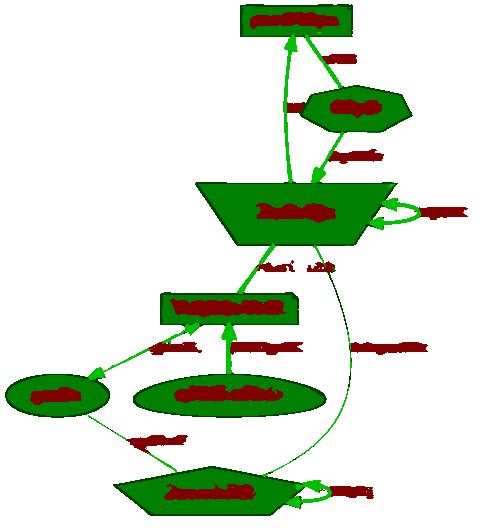}
    \caption{Example segmented image from the test set, showing the performance of the semantic segmentation algorithm on a clean full-size instance rather than a patch.}
    \label{fig:example_whole}
\end{figure}

In the case of semantic segmentation using bottom-up pixel classification, as our model does, a classification method learned on patches can be readily applied to whole images. Since the patches used are drawn uniformly from the complete images of diagrams in our dataset, the results on patch-level data are expected to be comparable to the results on complete images provided that edge effects do not interfere. As Figure \ref{fig:example_whole} shows, the results on complete images are consistent with the results on patches.

\subsection{Variant Models}
We also tested variations on the model using the augmented dataset to determine the effects of reducing the size of the model. While the quantitative performance is similar for reduced models, as shown in Table \ref{tab:results_aug}, the full model has the best performance in terms of pixel classification accuracy. The qualitative nature of the errors, as shown in Figure \ref{fig:examples_full_patches_aug}, differs significantly between the models. As expected, the coarse branch used in isolation provides a less precise segmentation, with broader regions rather than fine lines. Without the use of the coarse branch, fine details are visible, but there are noticeably more errors, especially around the borders of regions; this effect is most noticeable in the case of the network in which the fine branch has been reduced to two $3 \times 3$ convolutional layers before the final classification module. Of the models tested, the full model provides the best performance both qualitatively and quantitatively.

\begin{table}
    \centering
    \caption{Summary of models and results using the augmented dataset.}
    \begin{tabular}{|c|r|c|}
        \hline
        \textbf{Model} & \textbf{Parameters} & \textbf{Accuracy} \\
        \hline
        Full & 450117 & 0.937 \\
        Coarse only & 233701 & 0.849 \\
        Fine only (8 convolutions) & 233701 & 0.907 \\
        Fine only (4 convolutions) & 85989 & 0.913 \\
        Fine only (2 convolutions) & 26469 & 0.888 \\
        \hline
    \end{tabular}
    \label{tab:results_aug}
\end{table}

\subsection{Clean Images}
\begin{figure}
    \centering
    \begin{tabular}{ccc}
         Image & Ground Truth & Segmented \\
         \frame{\includegraphics[width=1.0in]{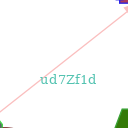}} & \frame{\includegraphics[width=1.0in]{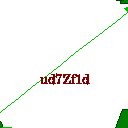}} & \frame{\includegraphics[width=1.0in]{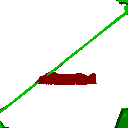}} \\
         \frame{\includegraphics[width=1.0in]{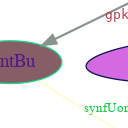}} & \frame{\includegraphics[width=1.0in]{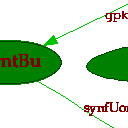}} & \frame{\includegraphics[width=1.0in]{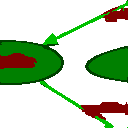}} \\
         \frame{\includegraphics[width=1.0in]{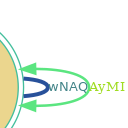}} & \frame{\includegraphics[width=1.0in]{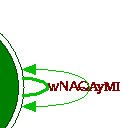}} & \frame{\includegraphics[width=1.0in]{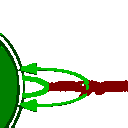}} \\
    \end{tabular}
    \caption{Examples of segmented patches, 128-by-128 pixels, from the test set.}
    \label{fig:examples_full_patches}
\end{figure}

The performance is similarly high when trained and tested exclusively on clean data without augmentation from JPEG-compressed images. The non-augmented dataset is a subset of the augmented dataset, consisting of the 2500 clean training images, 500 clean validation patches, and 500 clean test patches. We used the same learning rate as in the tests with augmented data: $10^{-6}$. The full semantic segmentation model achieves accuracy of 93.6\% on patches from the validation set and 93.7\% on the patches from the test set in this scenario, which is very similar to the performance on augmented data. Figure \ref{fig:examples_full_patches} shows example patches from the clean dataset. A few misclassified pixels are visible, notably in the bottom row, where some edge pixels are classified as node borders and \textit{vice versa}.

\section{Discussion}
\label{sec:discussion}
Our results demonstrate very high performance of our model at a range of scales for semantic segmentation of node-and-edge diagrams.

In order to support an assistive interface like TADA, it will be necessary to perform instance segmentation as well as semantic segmentation. In this case, the use of a pipeline consisting of semantic segmentation followed by instance segmentation is logical, as nodes and text can be expected to be well-separated in well-designed diagrams; as a result, connected components in the semantic segmentation will provide a starting point for the instance segmentation of these classes. Edges will frequently cross each other, which makes the use of connected components as a basis for instance segmentation more challenging for this class.

While dedicated node-link diagram assistive interfaces require additional features from a diagram parsing system, the semantic segmentation approach is already applicable to some other existing approaches to making diagrams more accessible. A semantic segmentation can, for example, be used to create bitmap components that can be later 3D printed or otherwise turned into tactile components (\textit{e.g.}, vector-based tactile graphics, or to supported existing research on automating tactile graphics creation as seen in the Tactile Graphics Project~\cite{tactileGraphicsProj}). The Chart4Blind assistive interface \cite{moured2024chart4blind}, which is designed for line graphs, uses instance segmentation and text recognition, as well as calibration of numerical values based on chart axes, to assist in generating accessible versions of diagrams which can be presented using tactile interfaces and Braille printers. While semantic segmentation alone is not sufficient to support Chart4Blind, and it is intended for a different class of graphical data, it shows the potential of this type of tactile presentation of diagrams.

\subsection{Future Work}
The model described in this paper shows excellent quantitative and qualitative performance on our dataset; any remaining improvements would be of marginal benefit on similar data. There are, however, related diagrams that are not addressed in our dataset. Pure line drawings, with no color cues to distinguish node interiors from background, are a special case on which our method may struggle, if the nodes are sufficiently large. Similarly, pure line drawings which use more complex symbols for nodes (\textit{e.g.}, circuit diagrams or engineering diagrams) are worth investigating further, as it would be useful to extend a very general segmentation model such as ours to handle additional classes to improve accessibility in these domains.

The addition of a probabilistic graphical model module, such as a dense conditional random field (CRF) \cite{zheng2015conditional}, is of interest as a way to incorporate broader context into our instance segmentation process. This may provide marginal improvements on our dataset, particularly for JPEG compressed images which have occasional ``stray'' pixels in the semantic segmentation, and holds promise for working with more challenging line drawing nodes.

Our work thus far has been in the context of images produced directly by rendering a diagram. It, however, also interesting to explore options for working with scanned or photographed diagrams. A diagram parsing pipeline capable of working with these images would be able to support assistive interfaces that can capture images of diagrams from the user's surroundings, parse them, and provide assistance in making use of the diagrams. This would require a substantially expanded dataset to capture the flaws introduced by the image capture process (either a photograph or a scan) as well as capturing the characteristics of rendered images, but is a promising line of research for extending the applicability of our model.

\section{Conclusion}
\label{sec:conclusion}
Our results show that accurate semantic segmentation of rendered node-link diagrams is feasible using compact deep learning models. Our models represent a complementary approach to existing methods for parsing node-link diagrams. This parsing problem is significant because node-link diagrams are a common method of conveying important information, and are not readily made accessible to blind and low-vision users without a method for parsing them. While existing assistive interfaces for this class of diagram require more than a semantic segmentation, a high-quality semantic segmentation is suitable to serve as a component of a parsing pipeline. Furthermore, given the quality of the semantic segmentations produced by our model, they may be suitable for supporting new assistive interfaces designed around their capabilities or for adapting node-link diagrams to interfaces originally designed for other domains. In addition to the quality of the segmentations produced by our models, the models are compact; this will make these models more suitable to running locally than large language models and multimodal large models, whose computational requirements bring significant disadvantages in terms of infrastructure requirements and environmental concerns.

\bibliographystyle{ieeetr}
\bibliography{references.bib}

\end{document}